\documentclass{article} % For LaTeX2e
\usepackage{iclr2026_conference,times}

\usepackage{amsmath,amsfonts,bm}

\def\eqref#1{equation~\ref{#1}}
\def\1{\bm{1}}

\DeclareMathAlphabet{\mathsfit}{\encodingdefault}{\sfdefault}{m}{sl}
\SetMathAlphabet{\mathsfit}{bold}{\encodingdefault}{\sfdefault}{bx}{n}

\usepackage{hyperref}
\usepackage{url}
\usepackage{rotating}
\usepackage{booktabs}       % professional-quality tables
\usepackage{tabularx}
\usepackage{graphicx} % Required for \resizebox
\definecolor{darkgrey}{rgb}{0.53,0.53,0.53}
\definecolor{mygrey}{rgb}{0.9,0.9,0.9}
\definecolor{blue1}{RGB}{166, 206, 227}
\definecolor{blue2}{RGB}{28, 118, 179}
\definecolor{myorange}{RGB}{192,97,21}
\usepackage[most]{tcolorbox}
\usepackage{CJKutf8}
\newcommand{\cc}[1]{\begin{CJK*}{UTF8}{gbsn}#1\end{CJK*}}
\usepackage{hyperref}
\usepackage{multirow}

\title{OmniEduBench: A Comprehensive Chinese Benchmark for Evaluating Large Language Models in Education}

\author{
Min Zhang$^{1}$\thanks{Project leader. Min Zhang (mzhang@cs.ecnu.edu.cn) } \quad 
Hao Chen$^{1}$ \quad 
Hao Chen$^{1}$ \quad 
Wenqi Zhang$^{2}$ \quad 
Didi Zhu$^{3}$ \quad 
Xin Lin$^{1}$ \\[0.3em] 
\ \textbf{Bo Jiang}$^{1}$\thanks{Corresponding authors. Bo Jiang (bjiang@deit.ecnu.edu.cn) and Aimin Zhou (amzhou@cs.ecnu.edu.cn)}  \quad \textbf{Aimin Zhou}$^{1\dagger}$ \quad 
\textbf{Fei Wu}$^{2}$ \quad 
\textbf{Kun Kuang}$^{2}$ \\[0.3em]
$^{1}$East China Normal University \quad 
$^{2}$Zhejiang University \quad 
$^{3}$Imperial College London \\[0.3em]
{\tt\small mzhang@cs.ecnu.edu.cn} \quad
{\tt\small bjiang@deit.ecnu.edu.cn} \quad
{\tt\small amzhou@cs.ecnu.edu.cn} \\[0.5em]
\quad \quad \quad \quad \quad
\textcolor{magenta}{\url{https://mind-lab-ecnu.github.io/OmniEduBench/}}
}

\iclrfinalcopy

\begin{document}

\maketitle

\begin{abstract}
    With the rapid development of large language models (LLMs), various LLM-based works have been widely applied in educational fields. However, most existing LLMs and their benchmarks focus primarily on \textit{the knowledge dimension, largely neglecting the evaluation of cultivation capabilities} that are essential for real-world educational scenarios. Additionally, current benchmarks are often \textit{limited to a single subject or question type, lacking sufficient diversity}. This issue is particularly prominent within the Chinese context. To address this gap, we introduce \textbf{OmniEduBench, a comprehensive Chinese educational benchmark}. OmniEduBench consists of 24.602K high-quality question-answer pairs. The data is meticulously divided into two core dimensions: \textbf{the knowledge dimension and the cultivation dimension}, which contain 18.121K and 6.481K entries, respectively. Each dimension is further subdivided into 6 fine-grained categories, covering a total of 61 different subjects (41 in the knowledge and 20 in the cultivation). Furthermore, the dataset features a rich variety of question formats, including 11 common exam question types, providing a solid foundation for comprehensively evaluating LLMs' capabilities in education. Extensive experiments on 11 mainstream open-source and closed-source LLMs reveal a clear performance gap. In the knowledge dimension, only Gemini-2.5 Pro surpassed 60\% accuracy, while in the cultivation dimension, the best-performing model, QWQ, still trailed human intelligence by nearly 30\%. These results highlight the substantial room for improvement and underscore the challenges of applying LLMs in education.
\end{abstract}

\section{Introduction}

With the rapid emergence of large language models (LLMs), evaluation benchmarks have become increasingly critical, shifting the focus of assessment toward broader and complex skills. To address the demands of this complex paradigm, a variety of benchmarks have been proposed to evaluate the diverse capabilities of LLMs. These benchmarks cover a wide spectrum of areas, including knowledge and language understanding (\textit{e.g.}, MMLU~\citep{hendrycks2021measuring}, ARC~\citep{allenai:arc}), reasoning (\textit{e.g.}, GSM8K~\citep{cobbe2021gsm8k}, AIME~\citep{patel2024aime}), multi-turn open-ended dialogue (\textit{e.g.}, MT-bench~\citep{bai2024mt}), and coding (\textit{e.g.}, MBPP~\citep{austin2021program}). Serving as indispensable tools for advancing LLM development, these benchmarks have been widely adopted in recent influential works~\citep{hurst2024gpt-4o,liu2024deepseek-v2,seed2025seed-oss,comanici2025gemini,taylor2022galactica,touvron2023llama,openai2023gpt4,hoffmann2022training}.

In recent years, a series of powerful Chinese LLMs emerged, such as the Qwen~\citep{qwen2.5,yang2025qwen3}, DeepSeek~\citep{liu2024deepseek-v2,guo2025deepseek,liu2024deepseek-v3}, achieving performance levels comparable to overseas LLMs. With the growing application of LLMs in education, researchers have also begun to propose Chinese education benchmarks, which can be broadly categorized into two types: \textcolor{blue2}{(1) datasets translated from other languages} and \textcolor{myorange}{(2) datasets natively constructed from Chinese education corpora}. Specifically, \textcolor{blue2}{(1) Datasets translated from other languages} refer to benchmarks constructed by directly translating existing benchmarks from other languages into Chinese. A representative work is CLUE~\citep{xu-etal-2020-clue}, which was translated from the English GLUE~\citep{wangglue}.
However, a simple translation approach is insufficient for a rigorous evaluation of LLMs in Chinese. These datasets often fail to reflect the unique linguistic and cultural challenges of the Chinese education and inherently carry biases from their original environment, thus limiting their ability to assess LLMs’ understanding of local education knowledge and teacher-student needs.

\textcolor{myorange}{(2) Datasets natively constructed from Chinese educational corpora} refer to benchmarks directly collected from Chinese educational text resources, such as C-Eval~\citep{huang2023ceval}, Edubench~\citep{xu2025edubench}, Scieval~\citep{sun2024scieval}, AGIEval~\citep{zhong2023agieval}, and SuperCLUE~\citep{xu2023superclue}. However, most existing education benchmarks are often limited to a single subject or question type, lacking sufficient diversity. Additionally, these datasets typically focus on the knowledge dimension, overlooking the unique cultivation aspects that are essential in real-world education. 

% This naturally raises a pivotal question:

% \begin{tcolorbox}[notitle, rounded corners, colframe=myorange, colback=white, boxrule=2pt, boxsep=0pt, left=0.15cm, right=0.17cm, enhanced, shadow={2.5pt}{-2.5pt}{0pt}{opacity=5,mygrey},toprule=2pt, before skip=0.65em, after skip=0.75em]
% \emph{
%   {
%     \centering 
%   {
%     \fontsize{8pt}{13.2pt}\selectfont 
%     How can we develop a natively Chinese evaluation benchmark that captures the unique linguistic and cultural knowledge of Chinese, incorporates diverse question types, and assesses large language models (LLMs) not only on the knowledge dimension but also on the distinctive Cultivation competencies required in realistic educational settings?
%   }
%   \\
%   }
%   }
% \end{tcolorbox}

\begin{figure}[tbp]
    \centering
    \includegraphics[height=0.6\textwidth]{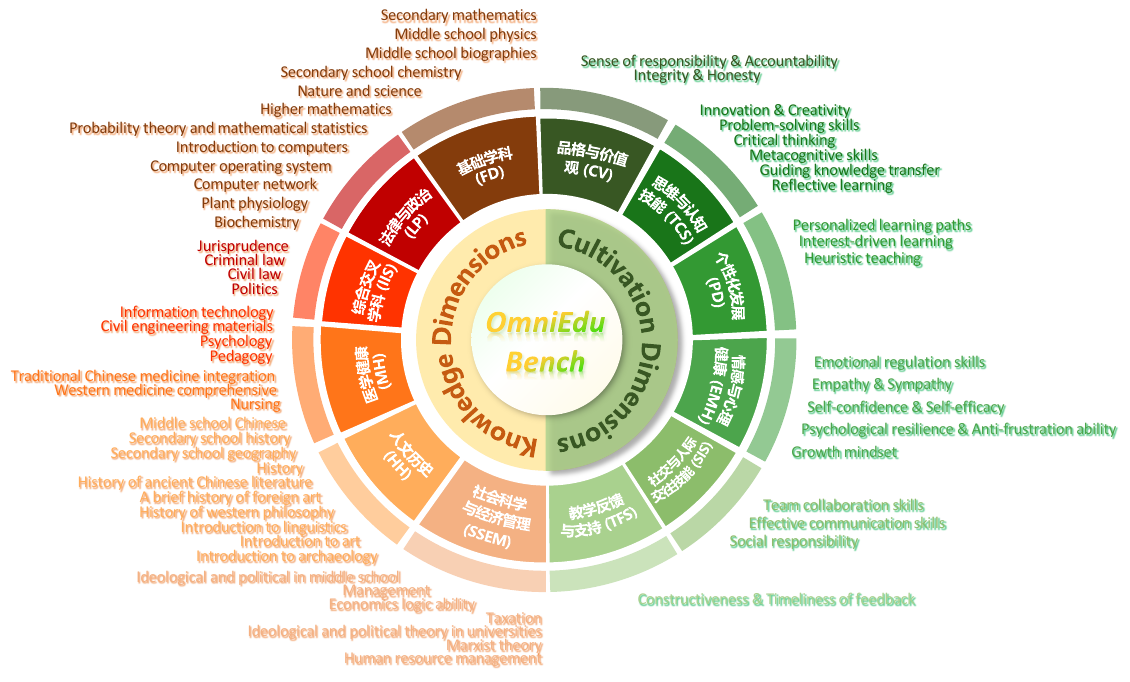}
    \vspace{-8mm}
    \caption{Overview of OmniEduBench. The benchmark comprises two dimensions: 41 subjects across six categories in the knowledge, and 20 subjects across six categories in the cultivation.}
    \label{fig:omniframe}
    \vspace{-6mm}
\end{figure}

We present OmniEduBench, a comprehensive Chinese education benchmark designed to thoroughly evaluate LLMs in terms of both knowledge understanding and skill cultivation in educational scenarios. OmniEduBench encompasses knowledge and cultivation dimension and comprises a total of 24.602K high-quality question–answer pairs, covering 11 common exam question types (\textit{e.g.}, multiple choice (\cc{单选题}), multiple answer (\cc{多选题}), fill-in-the-blank (\cc{填空题}), short answer (\cc{简答题}), composite questions (\cc{复合题}), term explanation (\cc{名词解释}), True/False (\cc{判断题}), calculation (\cc{计算题}), logical reasoning (\cc{逻辑推理题}), case analysis (\cc{案例分析题}), and essay (\cc{论述题})), as illustrated in Figure~\ref{fig:omniframe}. The knowledge dimension includes 18.121K question–answer pairs spanning 41 subject areas, from humanities to science and engineering, and covering five difficulty levels: elementary school, middle school, high school, college, and professional examinations. The cultivation dimension comprises 6.481K question–answer pairs across 20 teaching-related comments, including guided teaching, student emotional support, and moral education (see Table~\ref{tab:sta} for details), aiming to comprehensively assess the diverse competencies required in real-world educational settings. Extensive experiments demonstrate that our proposed OmniEduBench presents a highly challenging and significant benchmark for Chinese educational evaluation. Additionally, we introduce OmniEduBench HARD, a high-difficulty subset of OmniEduBench, specifically targeting particularly demanding subjects such as advanced mathematics and competitions that require sophisticated reasoning skills. Even the state-of-the-art LLMs achieve less than 50\% accuracy on this subset, highlighting the rigor and necessity of our proposed OmniEduBench education benchmark.

\vspace{-3mm}
\section{OmniEduBench}

Our proposed OmniEduBench education benchmark is designed as a \textcolor{myorange}{natively Chinese education evaluation benchmark} that captures the unique linguistic and cultural knowledge of Chinese education, encompasses diverse question types, and assesses LLMs not only on their knowledge capabilities but also on the distinctive cultivation competencies required in real-world educational scenarios. 

\subsection{Task Definition}

\textbf{Knowledge dimension} focuses on evaluating the model’s mastery of subject-specific knowledge. Tasks in this dimension include 11 common exam question types (\textit{e.g.}, multiple choice, multiple answer, fill-in-the-blank, short answer, composite questions, term explanation, True/False, calculation, logical reasoning, case analysis, and essay). These 11 question types span a wide range of disciplines, from humanities and history to science, engineering, and professional fields. The primary goal is to assess the LLM’s problem-solving capabilities within the context of real-world education. 

\textbf{Cultivation dimension} assesses LLMs on their ability to support holistic educational objectives beyond mere knowledge acquisition. This includes guiding students’ thinking processes, fostering moral and value development, enhancing emotional understanding, and promoting critical reasoning skills. Tasks in this dimension are designed to reflect realistic learning scenarios, where models must provide pedagogically sound feedback that aligns with students’ cognitive and emotional needs.

\begin{figure}[tbp]
    \centering
    \includegraphics[height=0.3\textwidth]{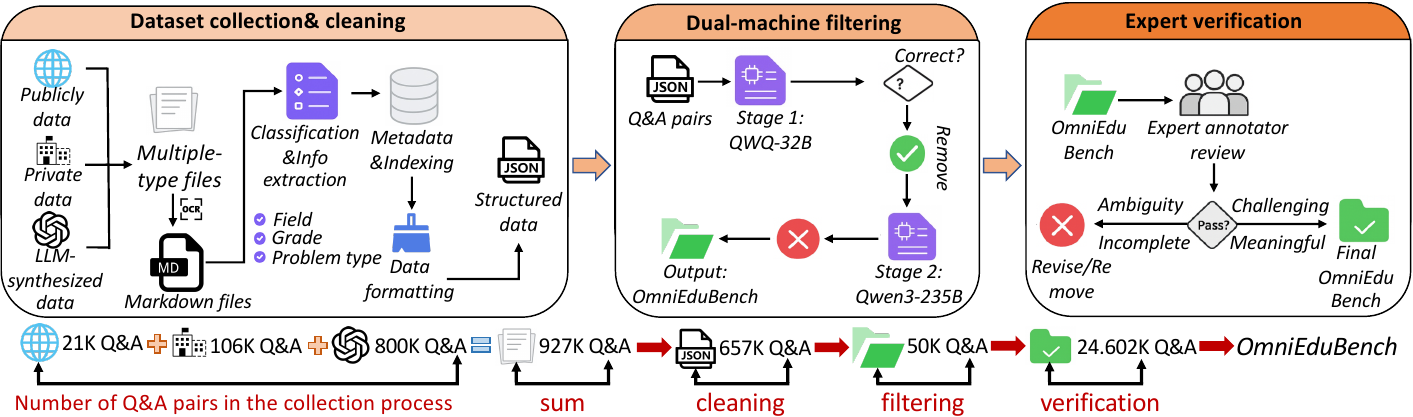}
    \vspace{-6mm}
    \caption{Overview of the construction process, including collection, cleaning, filtering, verification.}
    \label{fig:omniprocess}
    \vspace{-1mm}
\end{figure}

\subsection{Benchmark Construction}
\label{subsec:bc}

In this section, we provide a detailed overview of the construction process for the proposed OmniEduBench education evaluation benchmark, as illustrated in Figure~\ref{fig:omniprocess}. The process consists of four key stages: dataset collection, dataset cleaning, dual-machine filtering, and expert verification.

\textbf{Dataset collection.} OmniEduBench is designed to encompass a wide range of diverse scenarios to enable comprehensive evaluation. To achieve this, we employ three distinct data collection methods, carefully balancing diversity and efficiency in the construction of the OmniEduBench benchmark.

\textit{Manual collection of publicly available data.} 
Existing benchmarks often lack sufficient diversity in question types and knowledge coverage, making them inadequate for our 41 subjects in knowledge dimensions. To address this gap, we manually collected additional data from publicly available online resources (\textit{e.g.}, XuekeNet, ZujuanNet, ShijuanNet, ShitiNet) to enrich diversity and ensure coverage of underrepresented scenarios, such as primary and career education. Furthermore, guided by the Catalogue of Undergraduate Programs in Regular Higher Education Institutions~\footnote{\url{http://www.moe.gov.cn/srcsite/A08/moe_1034/s4930/202403/W020240319305498791768.pdf}} issued by China’s Ministry of Education, we curated a large body of review materials and exam questions across 13 academic disciplines, including philosophy, education, law, literature, history, science, engineering, agriculture, medicine, military science, management, and the arts. This effort significantly improves distributional balance and provides a more faithful reflection of real applications.

\textit{Manual collection of private data.}
Data contamination remains one of the most critical challenges in constructing evaluation datasets for LLMs. To mitigate this risk, we manually collected additional data from private resources, such as internal school exam papers. Unlike widely circulated national exams, these materials have never appeared on the public Internet or been included in large-scale web crawls, effectively reducing the risk of leakage. Incorporating such private data enhances the reliability and fairness of the benchmark, while providing a more rigorous assessment of models.

\textit{LLM-generated data.}
Given the difficulty of directly obtaining data in the cultivation dimension, we leveraged LLMs to generate a substantial number of scenario-based question–answer pairs, aiming to supplement gaps in existing resources. To ensure the quality of the synthetic data, we invited five education experts to conduct discussions on 20 cultivation subjects and consulted relevant books, papers, and other materials. The collected content was organized into a database, which was then provided to the LLM to enhance the fidelity and accuracy of the generated data. For generated questions, to increase their challenge, we designed highly confounding distractors via prompts and conducted sampling checks and revisions with expert verification (please see more details in expert verification). 
\textcolor{myorange}{Finally, we collected a total of 927K question–answer (Q\&A) pairs, including 21K from publicly available data, 106K from private data, and 800K generated by LLMs.}

\textbf{Dataset cleaning.}
The entire data cleaning process consists of multiple steps. (1) We used MinerU~\citep{wang2024mineru} to convert the collected 927K Q\&A pairs into Markdown (md) format, enabling structured management and efficient information extraction. (2) Detailed metadata were extracted for each question, including subject, grade level, question type, and knowledge tags, to construct comprehensive question profiles that facilitate data management and subsequent analysis. (3) Standard data cleaning procedures were applied, including deduplication, removal of questions with missing key content, filtering of sensitive or inappropriate content, and exclusion of questions that rely on external information.\textcolor{myorange}{After the cleaning process, we obtained a total of 657K Q\&A pairs.}

\textbf{Dual-machine filtering.}
To ensure OmniEduBench is a high-quality and challenging benchmark, we implemented a dual-model filtering mechanism on an \textcolor{myorange}{initial set of 657K Q\&A pairs}. Specifically, we first evaluated all questions using QWQ32B~\citep{qwq32b}, retaining only those that the model answered incorrectly. This initial filtering resulted in \textcolor{myorange}{a subset of 430K Q\&A pairs}. These questions then underwent a second filtering stage with the same strategy, this time using Qwen3-235B~\citep{yang2025qwen3}, ultimately yielding the final \textcolor{myorange}{set of 50K high-quality} and challenging data. 

\textbf{Expert verification.}
We recruited 50 master's students to perform an initial quality check on the dataset based on five predefined dimensions (as shown in Table~\ref{tab:ev}), removing any data that did not meet the criteria, which resulted in \textcolor{myorange}{a final set of 24.602K Q\&A pairs} for OmniEduBench. Subsequently, we invited 5 senior annotation experts to conduct a rigorous quality review on a 15\% random sample of the OmniEduBench. The review results, shown in Table~\ref{tab:ev}, indicate that the dataset maintains high overall quality, demonstrating both reliability and applicability.

\subsection{Evaluation Criteria}
\label{subsec:ec}

Based on the characteristics of different question types, we adopt two evaluation metrics: (1) \textbf{Choice}. For questions with a standard answer, we directly evaluate the provided answer. This simplifies the scoring process, as the model only needs to select the most appropriate option, thereby reducing ambiguity in assessment. (2) \textbf{LLM-assisted scoring}. For short-answer questions that may have multiple valid forms but are semantically equivalent, we employ an LLM-assisted scoring method. This approach provides greater flexibility, avoids imposing unnecessary constraints on the model, and allows for a more accurate evaluation of the model’s semantic understanding and expression.

\begin{table}[tbp]
    \centering
    \caption{Statistics of OmniEduBench and more detailed per-subject information are shown in the Appendix. Bilingual names and abbreviations of six knowledge and six cultivation dimensions.}
    \vspace{0.2mm}
    \resizebox{0.99\textwidth}{!}{
    \begin{tabular}{lcc|lcc}
        \toprule
        \multicolumn{3}{c|}{\textcolor{myorange}{\textit{Knowledge dimension}}} & \multicolumn{3}{c}{\textcolor{myorange}{\textit{Cultivation dimension}}}\\
        English name & Abbreviation & Chinese name & English name & Abbreviation & Chinese name \\
        \midrule
        Law \& Politics  & LP &  \cc{法律与政治} & Character \& Values & CV & \cc{品格与价值观} \\ 
        Foundational Disciplines & FD & \cc{基础学科} & Personalized Development & PD & \cc{个性化发展} \\
        Humanities \& History & HH & \cc{人文与历史} & Social \& Interpersonal Skills & SIS & \cc{社会与人际交往} \\
        Medicine \& Health & MH & \cc{医学与健康}  & Thinking \& Cognitive Skills & TCS & \cc{思维与认知能力} \\
        Interdisciplinary \& Integrated Subjects & IIS & \cc{综合与交叉学科} &  Teaching Feedback \& Support & TFS & \cc{教学反馈与支持}) \\
        Social Sciences \& Economics Management & SSEM & \cc{社会科学与经济管理} & Emotional \& Mental Health & EMH & \cc{情感与心理健康} \\
        \bottomrule
    \end{tabular}}
%     \label{tab:name}
%     \vspace{-6mm}
% \end{table}
% \begin{table}[tbp]
%     \centering
%     \caption{Statistics of OmniEduBench with detailed per-subject information shown in the Appendix.}
%     \vspace{0.2mm}
    \resizebox{0.99\textwidth}{!}{
    \begin{tabular}{lcc|lcc|lcc}
        % \toprule
        Category & Subjects & Questions & Category & Subjects & Questions & Category & Subjects & Questions \\
        \midrule
        \multicolumn{3}{c|}{\textcolor{myorange}{\textit{In terms of dimension}}} & \multicolumn{3}{c|}{\textcolor{myorange}{\textit{In terms of Knowledge}}} & \multicolumn{3}{c}{\textcolor{myorange}{\textit{In terms of Cultivation}}} \\
        Knowledge  & 41 & 18,121 & LP & 4 & 1,455 & CV & 2 & 694 \\
        Cultivation & 20 & 6,481 & FD  & 11 & 7,918 & PD & 3 & 1,031 \\
        \multicolumn{3}{c|}{\textcolor{myorange}{\textit{In terms of different level}}} & HH & 10 & 5,331 & SIS & 3 & 736 \\
        K-12 Schools  & 10 & 4,384 & MH & 3 & 918 & TCS & 6 & 1,900 \\
        High school & 11 & 6,735 & IIS & 4 & 914 & TFS & 1 & 193 \\
        College  & 30 & 6,364 & SSEM & 9 & 1,643 & EMH & 5 & 1,833 \\
        \midrule
        Total  & 61 & 24,602 & Total  & 41 & 18,179 & Total  & 20 & 6,387 \\
        \bottomrule
    \end{tabular}}
    \label{tab:sta}
    \vspace{-4mm}
\end{table}

\begin{table}[tbp]
    \centering
    \caption{Expert validation results for the OmniEduBench dataset.}
    \vspace{0.2mm}
    \resizebox{0.99\textwidth}{!}{
    \begin{tabular}{lcccc}
        \toprule
        Metric English name & Metric Chinese name & Average & Standard deviation & Inter-rater agreement  \\
        \midrule
        Overall quality & \cc{整体质量} & 4.8 & 0.1 & 0.90\\
        Clarity & \cc{问题清晰度} & 4.5 & 0.2 & 0.85 \\
        Option perplexity & \cc{选项困惑度} & 4.8 & 0.3 & 0.83 \\
        Accuracy & \cc{答案准确性} & 4.8 & 0.1 & 0.90 \\
        Cultivation value & \cc{育人价值} & 4.6 & 0.2 & 0.88 \\
        \bottomrule
    \end{tabular}}
    \label{tab:ev}
    \vspace{-4mm}
\end{table}

\subsection{Statistics}

\begin{figure}[tbp]
    \centering
    \includegraphics[height=0.32\textwidth]{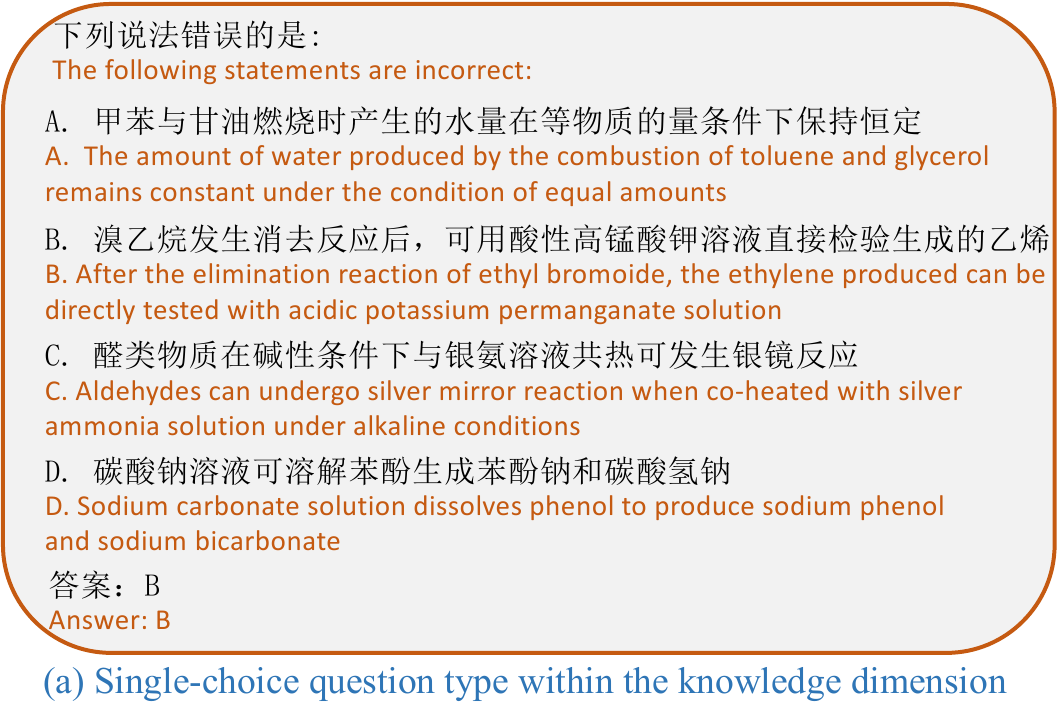}
    \includegraphics[height=0.32\textwidth]{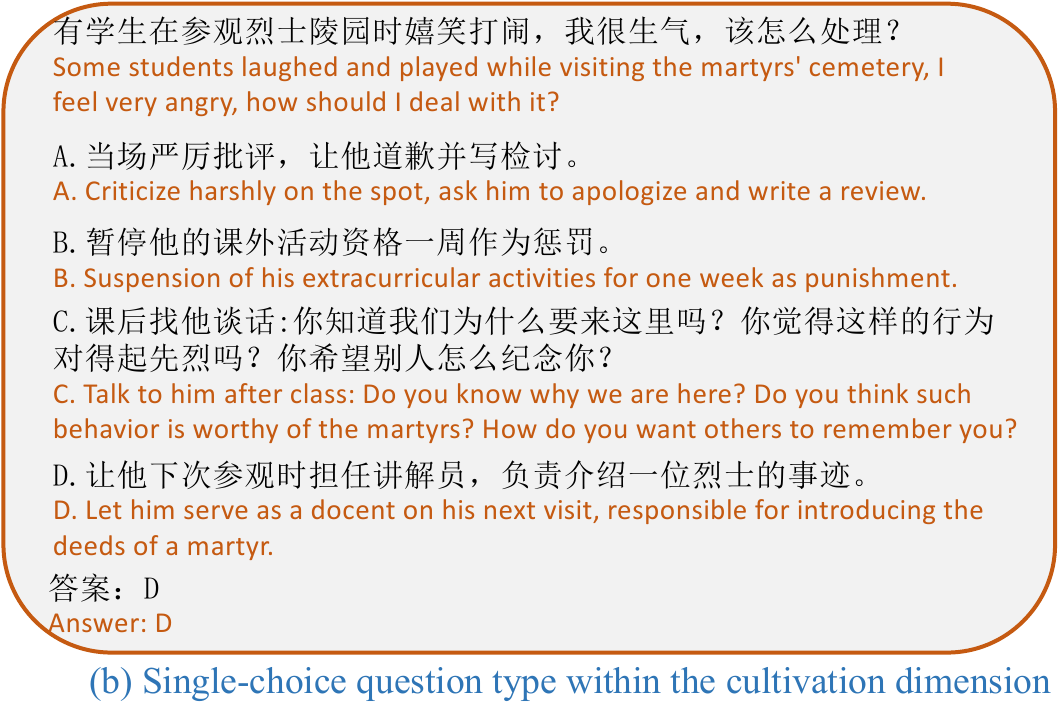}
    \vspace{-3mm}
    \caption{Example of (a) a single-choice question in the knowledge from a college chemist. (b) A single-choice question in the cultivation. English translations are shown for better readability.}
    \label{fig:omni12}
    \vspace{-3mm}
\end{figure}

\begin{figure}[tbp]
    \centering
    \includegraphics[height=0.3\textwidth]{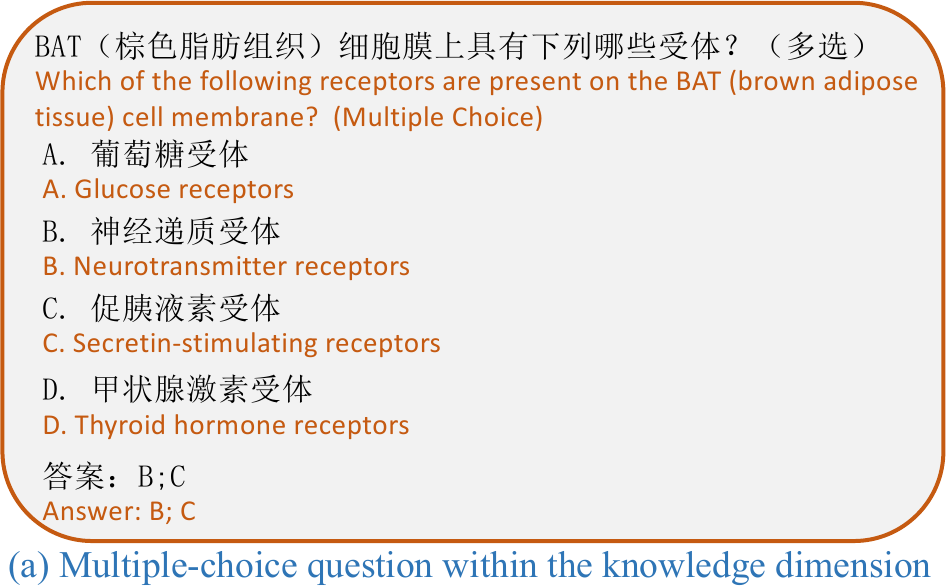}
    \includegraphics[height=0.3\textwidth]{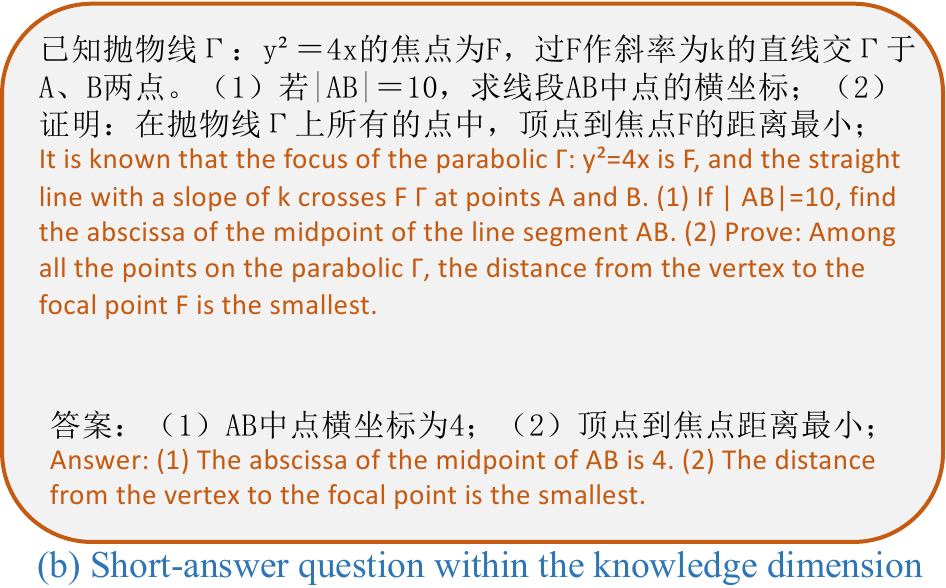}
    \vspace{-3mm}
    \caption{Example of (a) a multiple-choice question in the knowledge from Biology. (b) A short-answer question in the knowledge from Math. English translations are shown for better readability.}
    \label{fig:omni34}
    \vspace{-3mm}
\end{figure}

Through rigorous data filtering and expert validation, we collected 18.121K high-quality question–answer pairs for the knowledge and 6.481K for the cultivation. As illustrated in Figure~\ref{fig:omniframe} and summarized in Table~\ref{tab:sta}, with more detailed per-subject statistics provided in the Appendix, the dataset spans 12 major categories, as shown in Table~\ref{tab:sta}, including K-12, higher school, university-level courses, and cultivation aspects such as emotion and reasoning, covering a total of 61 specific scenarios. Figures~\ref{fig:omni12} and~\ref{fig:omni34} present some representative examples in different dimensions and question types. The questions exhibit wide variability in type and difficulty and are sourced from diverse origins, primarily newly collected from public or private resources or manually constructed. 

\section{Experiments}

In this section, we evaluate the performance of state-of-the-art (SOTA) methods in both English and Chinese. The experimental results indicate that OmniEduBench remains a competitive benchmark.

\subsection{Experimental Setup}

\begin{table}[tbp]
    \centering
    \caption{Zero-shot average accuracy (\%) across six categories in the \textbf{knowledge}. The highest accuracy is \textbf{bold}, and the second highest is \underline{underlined}. More results are provided in the Appendix.}
    \vspace{0.2mm}
    \resizebox{0.99\textwidth}{!}{
        \begin{tabular}{lccc|cccccc|c}
            \toprule
            \textbf{Model} & \textbf{Parameters} & \textbf{Access} & \textbf{Creator} & \textbf{FD} & \textbf{HH} & \textbf{SSEM} & \textbf{LP} & \textbf{MH} & \textbf{IIS} & \textbf{Average} \\ \midrule
            Qwen3  & 8B & Weights & Alibaba & 53.02 & 38.53 & 36.58 & 30.17 & 36.71 & 37.75 & 43.86 \\
            Qwen3   & 14B & Weights & Alibaba & 36.32 & 36.78 & 35.12 & 27.29 & 36.82 & 35.67 & 35.62 \\
            MuduoLLM & 14B & Weights & BNU \& TAL & 28.20 & 40.82 & 32.99 & 36.15 & 39.11 & 31.40 & 33.68 \\
            QwQ   & 32B & Weights & Alibaba & \underline{61.25} & 48.51 & 42.24 & \underline{49.90} & 55.01 & 47.26 & \underline{53.87} \\
            Seed-OSS     & 36B & Weights & ByteDance & 48.81 & \underline{50.14} & \underline{45.34} & 48.66 & \textbf{61.00} & \underline{49.56} & 49.53 \\
            Qwen2.5 & 72B & Weights & Alibaba & 19.53 & 30.95 & 20.57 & 13.26 & 23.86 & 20.90 & 22.76 \\
            Qwen3   & 235B (22B active) & Weights & Alibaba & 34.24 & 47.01 & 36.21 & 44.26 & 58.71 & 46.61 & 40.82 \\
            DeepSeek-V3.1 & 671B (37B active) & Weights & DeepSeek & 31.65 & 40.65 & 35.00 & 29.42 & 50.54 & 45.19 & 36.05 \\
            \midrule
            GPT-4o  & Undisclosed & API & OpenAI & 21.15 & 26.94 & 23.92 & 22.13 & 34.75 & 27.13 & 24.17 \\
            Claude-4 Sonnet  & Undisclosed & API & Anthropic & 41.49 & 44.29 & 35.36 & 27.56 & 34.86 & 42.34 & 40.35 \\
            Gemini-2.5 Pro  & Undisclosed & API & Google & \textbf{73.83} & \textbf{55.13} & \textbf{46.68} & \textbf{55.40} & \underline{60.68} & \textbf{54.16} & \textbf{62.76} \\
            \bottomrule
        \end{tabular}}
    \label{tab:main_results_kd}
    \vspace{-5.6mm}
\end{table}

\textbf{Baselines.}
We evaluate 11 mainstream large language models (LLMs) in total, including 3 cutting-edge closed-source models and 8 open-source models, one of which is a newly released education-oriented model. The closed-source models are GPT-4o~\citep{hurst2024gpt-4o}, Gemini-2.5 Pro~\citep{comanici2025gemini}, and Claude-4 Sonnet~\citep{claude4sonnet}. For the open-source models, we consider two main factors. First, they are grouped by parameter size into small (8B), medium (14B/32B/36B), and large (72B/235B/671B) scales. Second, they are categorized by functionality into: (a) general instruction-following models (Qwen2.5~\citep{qwen2.5}, Qwen3~\citep{yang2025qwen3}); (b) general reasoning models (QwQ~\citep{qwq32b}, Seed-OSS~\citep{seed2025seed-oss}, DeepSeek-V3.1~\citep{liu2024deepseek-v3}); and (c) education-specific models (MuduoLLM~\citep{muduollm2025}).

\textbf{Implementation details.}
In our experimental setup, we evaluate all large language models under both zero-shot and few-shot settings, with few-shot examples (0-, 1-, 3-, and 5-shot) drawn from a separately partitioned development set, distinct from the evaluation set. All open-source models are run using their official code, while closed-source models are accessed via official APIs. We consistently use Gemini-2.5 Pro as the LLM-assisted scoring model, unless otherwise specified.

\subsection{Main Results}

We evaluated all baseline models on OmniEduBench, reporting both per-task category and overall accuracy, as shown in Tables~\ref{tab:main_results_kd} and~\ref{tab:main_results_cd}). Results show that in the knowledge dimension, Gemini-2.5 Pro achieves the highest accuracy at 62.78\%, while in the cultivation dimension, the reasoning-enhanced version of QWQ performs best with an accuracy of 70.27\%. This performance highlights the challenging nature and strong discriminative power of the constructed OmniEduBench.

In the knowledge dimension, it is evident that, except for Gemini-2.5 Pro, closed-source models generally perform worse than open-source models on our OmniEduBench. For example, GPT-4o achieves an accuracy of 24.17\%, far below that of Qwen3-8B. This may indicate that the GPT series has relatively weak robustness when handling Chinese education exam-style questions. Meanwhile, model architecture has a significant impact on performance, such as Seed-OOS outperforms the Qwen family by more than 10\%. In the cultivation dimension, models generally perform better than in the knowledge dimension, which may be due to the fact that the cultivation tasks mainly consist of multiple-choice questions, making them simpler compared to knowledge tasks with 11 common exam question types. However, differences in performance between different model architectures still exist. Overall, GPT-40 performs the worst in both dimensions, with accuracy largely concentrated around 59.57\%, possibly because it has not been specifically optimized for this dimension.

\subsection{Analysis and Findings}

In this section, we further conduct extensive experiments at multiple levels, including few-shot examples, OmniEduBench HARD, and various LLM-assisted scoring methods.

\textbf{Results in few-shot examples.} In Table~\ref{tab:few-shot_kd}, we present in-context experimental results using different numbers of shots. As the number of shots increases, model performance generally improves; however, the overall gain is limited when considering the average results. We speculate that the drop in accuracy for some models is due to the fact that they have not (or not appropriately) incorporated few-shot examples during the instruction tuning stage. These findings suggest that while few-shot prompting can be beneficial for certain models, its effectiveness strongly depends on the model’s pretraining and instruction tuning strategies. Moreover, the limited average improvement indicates that simply increasing the number of shots may not always lead to substantial gains, highlighting the need for more sophisticated methods to integrate few-shot examples effectively. 

\begin{table}[tbp]
    \centering
    \caption{Zero-shot average accuracy (\%) across six categories in the \textbf{cultivation}. The highest accuracy is \textbf{bold}, and the second highest is \underline{underlined}. More results are provided in the Appendix.}
    \vspace{0.2mm}
    \resizebox{0.99\textwidth}{!}{
        \begin{tabular}{lccc|cccccc|c}
            \toprule
            \textbf{Model} & \textbf{Parameters} & \textbf{Access} & \textbf{Creator} & \textbf{TCS} & \textbf{EMH} & \textbf{SIS} & \textbf{CV} & \textbf{PD} & \textbf{TFS} & \textbf{Average} \\ \midrule
            Qwen3 & 8B & Weights & Alibaba  & 70.95 & 66.67 & 69.16 & 62.25 & 70.13 & \textbf{77.20} & 68.62 \\
            Qwen3  & 14B & Weights & Alibaba & 67.79 & 60.77 & 63.72 & 56.20 & 64.31 & 71.50 & 63.60 \\
            MuduoLLM  & 14B & Weights & BNU \& TAL & 64.42 & 60.77 & 63.45 & \textbf{66.14} & 67.51 & 64.77 & 63.96 \\
            QwQ & 32B &Weights & Alibaba  & \textbf{73.16} & \underline{68.36} & 69.84 & 65.13 & \textbf{71.77} & \underline{72.02} & \textbf{70.27} \\
            Seed\mbox{-}OSS & 36B &Weights & ByteDance & 70.74 & 65.30 & 66.03 & 62.82 & 67.12 & 70.47 & 67.18 \\
            Qwen2.5 & 72B &Weights & Alibaba  & 67.89 & 64.38 & 65.62 & 59.51 & 65.57 & 67.88 & 65.34 \\
            Qwen3 & 235B (22B active) &Weights & Alibaba & 67.84 & 61.10 & 64.54 & 55.76 & 64.40 & 70.47 & 63.74 \\
            DeepSeek\mbox{-}V3.1  & 671B (37B active) &Weights& DeepSeek & 71.58 & 65.41 & 69.02 & 61.96 & \underline{71.00} & \textbf{77.20} & 68.55 \\
            \midrule
            GPT\mbox{-}4o & Undisclosed & API & OpenAI & 61.63 & 59.57 & 59.24 & 55.33 & 57.71 & 65.80 & 59.57 \\
            Claude\mbox{-}4\mbox{-}sonnet  & Undisclosed & API & Anthropic        & 71.95 & \textbf{70.05} & \textbf{70.92} & 64.55 & 69.25 & 71.50 & \underline{70.03} \\
            Gemini\mbox{-}2.5\mbox{-}pro & Undisclosed & API & Google      & \underline{72.26} & 66.07 & \underline{70.79} & \underline{65.71} & 70.32 & 67.36 & 69.14 \\
            \bottomrule
        \end{tabular}}
    \label{tab:main_results_cd}
    \vspace{-2mm}
\end{table}

\begin{table}[tbp]
    \centering
    \caption{Average accuracy (\%) across six categories in one-shot, three-shot, and five-shot settings for the knowledge dimension. The highest accuracy is \textbf{bold}, and the second highest is \underline{underlined}.}
    \vspace{0.2mm}
    \resizebox{0.99\textwidth}{!}{
        \begin{tabular}{lccc|cccccc|c}
            \toprule
            \textbf{Model} & \textbf{Parameters} & \textbf{Access} & \textbf{Creator} & \textbf{FD} & \textbf{HH} & \textbf{SSEM} & \textbf{LP} & \textbf{MH} & \textbf{IIS} & \textbf{Average} \\ \midrule
            \multicolumn{11}{c}{\textcolor{myorange}{\textit{One-shot setting}}} \\
            Qwen3 & 8B & Weights & Alibaba & \textbf{52.80} & 46.45 & \textbf{41.90} & 29.76 & 40.20 & 40.59 & \textbf{41.95} \\
            MuduoLLM & 14B & Weights & BNU \& TAL & 27.36 & \underline{47.79} & 36.72 & 34.98 & 40.74 & 34.35 & 36.99 \\
            Qwen2.5 & 72B & Weights & Alibaba & 21.42 & 40.43 & 27.04 & 20.96 & 28.10 & 22.65 & 26.77 \\
            Qwen3 & 235B (22B activate) & Weights & Alibaba & \underline{37.72} & \textbf{60.79} & \underline{44.03} & \textbf{45.77} & \textbf{59.59} & \textbf{54.05} & \textbf{50.12} \\
            DeepSeek-V3.1 & 671B (37B activate) & Weights & DeepSeek & 30.00 & 41.73 & 34.65 & 30.72 & \underline{49.67} & 42.12 & \underline{38.15} \\
            \midrule
            \multicolumn{11}{c}{\textcolor{myorange}{\textit{Three-shot setting}}} \\
            Qwen3 & 8B & Weights & Alibaba & \textbf{52.98} & 46.00 & \textbf{39.74} & 30.65 & 39.32 & 40.85 & \underline{41.59} \\
            MuduoLLM & 14B & Weights & BNU \& TAL & 27.32 & \underline{46.86} & 35.79 & 33.81 & 39.54 & 32.42 & 35.96 \\
            Qwen2.5 & 72B & Weights & Alibaba & 21.43 & 40.86 & 27.27 & 20.41 & 27.12 & 23.88 & 26.83 \\
            Qwen3 & 235B (22B activate) & Weights & Alibaba & \underline{37.52} & \textbf{60.70} & \underline{43.40} & \textbf{45.77} & \textbf{59.48} & \textbf{52.57} & \textbf{49.54} \\
            DeepSeek-V3.1 & 671B (37B activate) & Weights & DeepSeek & 29.09 & 41.42 & 34.02 & 28.59 & \underline{48.80} & \underline{42.06} & 37.33 \\
            \midrule
            \multicolumn{11}{c}{\textcolor{myorange}{\textit{Five-shot setting}}} \\
            Qwen3 & 8B & Weights & Alibaba & \textbf{56.86} & 46.70 & \textbf{39.44} & 30.65 & 38.24 & 42.23 & \underline{42.35} \\
            MuduoLLM & 14B & Weights & BNU \& TAL & 26.93 & \underline{46.57} & 36.28 & 35.74 & 39.11 & 34.57 & 36.53 \\
            Qwen2.5 & 72B & Weights & Alibaba & 21.46 & 41.19 & 26.96 & 20.82 & 26.03 & 28.56 & 27.50 \\
            Qwen3 & 235B (22B activate) & Weights & Alibaba & \underline{37.40} & \textbf{60.41} & \underline{44.13} & \textbf{45.77} & \textbf{58.61} & \textbf{55.58} & \textbf{50.32} \\
            DeepSeek-V3.1 & 671B (37B activate) & Weights & DeepSeek & 29.39 & 41.17 & 32.87 & 28.45 & \underline{47.49} & 38.95 & 36.39 \\
            \bottomrule
        \end{tabular}}
    \label{tab:few-shot_kd}
\end{table}

\begin{figure}[tbp]
    \centering
    \includegraphics[height=0.41\textwidth]{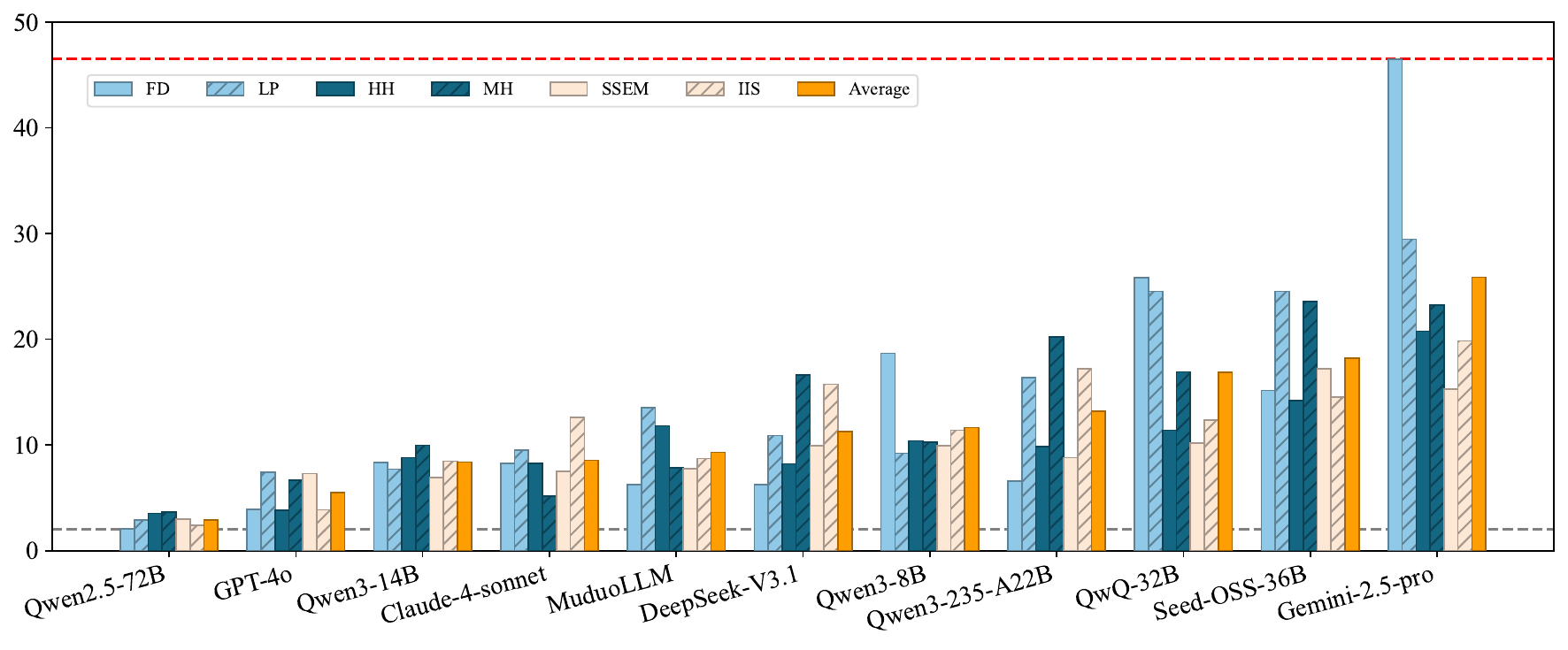}
    \vspace{-5mm}
    \caption{Zero-shot average accuracy (\%) on the knowledge  dimension of OmniEduBench HARD.}
    \label{fig:omnihard_kd}
    \vspace{-3mm}
\end{figure}

\begin{figure}[tbp]
    \centering
    \includegraphics[height=0.41\textwidth]{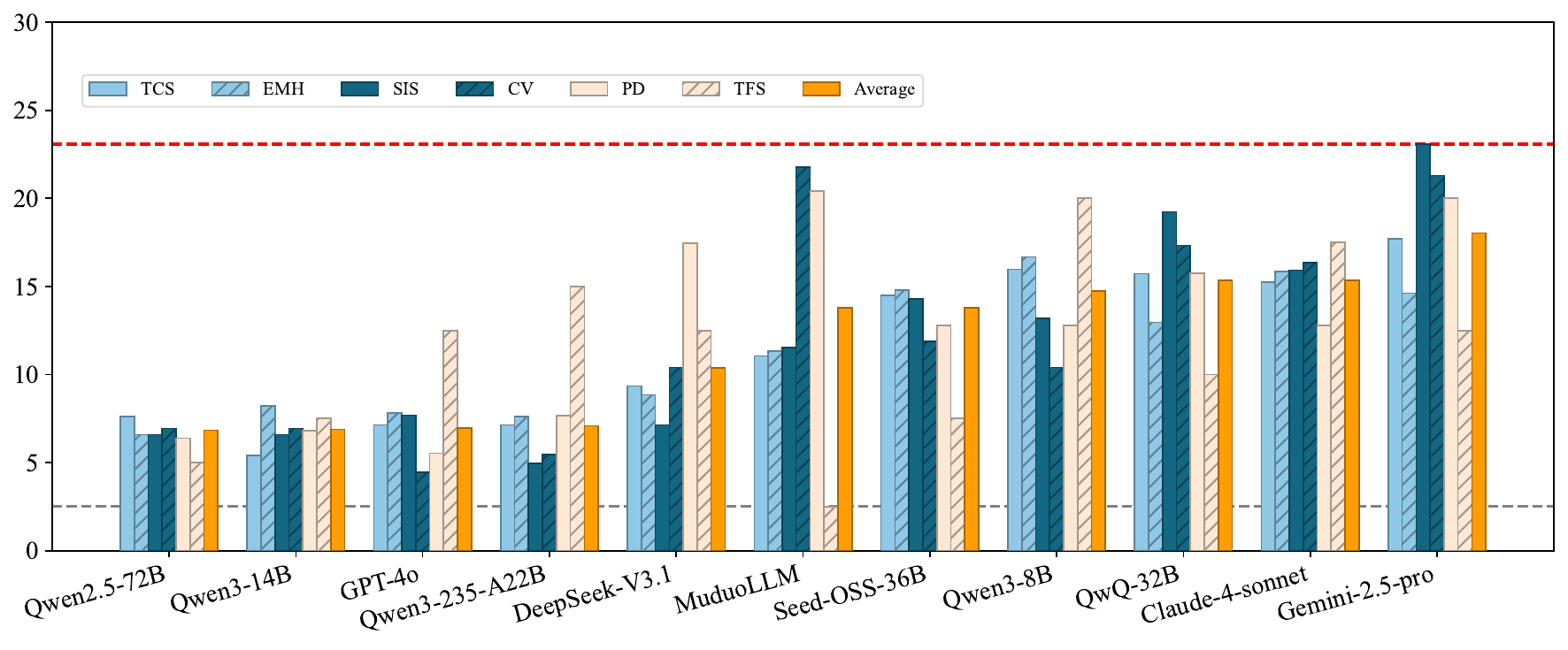}
    \vspace{-5mm}
    \caption{Zero-shot average accuracy (\%) on the cultivation  dimension of OmniEduBench HARD.}
    \label{fig:omnihard_cd}
    \vspace{-3mm}
\end{figure}

\textbf{Results on OmniEduBench HARD.} In Figures~\ref{fig:omnihard_kd} and~\ref{fig:omnihard_cd}, we present the average accuracy of each model on OmniEduBench HARD. OmniEduBench HARD is a subset of OmniEduBench, consisting of the bottom 26\% of samples based on model performance, including approximately 1.552K cultivation samples and 7.620K knowledge samples, for a total of 9.172K examples. The experimental results show that: (1) all 11 LLMs exhibit a significant performance drop on OmniEduBench HARD, with even the best-performing model, Gemini, achieving less than 50\% accuracy; (2) Qwen2.5-72 performs the worst, significantly lower than the other models, indicating limited capability in handling difficult samples. These findings indicate that further research is needed to enhance LLMs’ ability to generalize and maintain high performance on hard subsets of educational benchmarks.

\textbf{Results using different LLM-assisted scoring methods.}
In Table~\ref{tab:main_results_kd_llms}, we present the experimental results using different LLM-assisted scoring methods. The performance of the scoring model directly affects the evaluation outcomes: higher-quality scoring models provide more accurate assessments, leading to more precise measurements of the evaluated models’ capabilities. In this study, we employed three scoring models of varying quality. Overall, GPT-4o performed relatively poorly as a scoring model, failing to accurately evaluate the responses of LLMs. Consequently, the overall effectiveness of LLM-assisted evaluation is reduced when GPT-4o is used, highlighting the critical importance of selecting high-quality scoring models to ensure accurate and meaningful assessments. These findings suggest that the choice of scoring model can substantially influence the perceived performance of evaluated LLMs, and careful selection of scoring models is necessary.

\begin{table}[tbp]
    \centering
    \caption{Zero-shot average accuracy (\%) across six categories in the knowledge using different LLM-assisted scoring methods. The highest accuracy is \textbf{bold}, and the second highest is \underline{underlined}.}
    \vspace{0.2mm}
    \resizebox{0.99\textwidth}{!}{
        \begin{tabular}{lccc|cccccc|c}
            \toprule
            \textbf{Model} & \textbf{Parameters} & \textbf{Access} & \textbf{Creator} & \textbf{FD} & \textbf{HH} & \textbf{SSEM} & \textbf{LP} & \textbf{MH} & \textbf{IIS} & \textbf{Average} \\ \midrule
            \multicolumn{11}{c}{\textcolor{myorange}{\textit{Qwen3-A235B-assisted scoring method~\citep{yang2025qwen3}}}} \\
% Qwen3  & 8B & Weights & Alibaba & 55.45 & 48.51 & 42.54 & 30.86 & 38.13 & 44.42 & 48.85 \\
% Qwen3   & 14B & Weights & Alibaba & 37.82 & 48.70 & 40.60 & 29.07 & 38.45 & 41.14 & 40.76 \\
% MuduoLLM & 14B & Weights & BNU \& TAL & 27.78 & 43.22 & 34.08 & 36.01 & 39.00 & 29.21 & 34.18 \\
QwQ   & 32B & Weights& Alibaba & \underline{61.26} & 55.82 & 45.22 & \underline{50.10} & 55.66 & 48.36 & \underline{56.39} \\
Seed-OSS     & 36B &Weights & ByteDance & 51.16 & \underline{63.04} & \underline{51.55} & 50.38 & \textbf{62.31} & \underline{57.33} & 55.49 \\
% Qwen2.5 & 72B & Weights& Alibaba & 21.05 & 40.90 & 25.62 & 14.71 & 25.93 & 22.21 & 27.08 \\
% Qwen3   & 235B (22B active) & Weights & Alibaba & 37.11 & 59.20 & 42.97 & 45.77 & 61.00 & 53.61 & 46.85 \\
% DeepSeek-V3.1 & 671B (37B active) &Weights & DeepSeek & 29.94 & 40.47 & 33.72 & 29.42 & 50.33 & 41.36 & 34.93 \\
GPT-4o  & Undisclosed & API & OpenAI & 23.59 & 36.77 & 28.79 & 23.92 & 35.73 & 31.40 & 28.96 \\
Claude-4 Sonnet  & Undisclosed & API & Anthropic & 43.54 & 55.52 & 40.47 & 27.77 & 36.17 & 47.48 & 45.34 \\
Gemini-2.5 Pro  & Undisclosed & API & Google & \textbf{75.01} & \textbf{65.67} & \textbf{52.95} & \textbf{56.29} & \underline{61.44} & \textbf{61.49} & \textbf{67.41} \\
            \midrule
            \multicolumn{11}{c}{\textcolor{myorange}{\textit{Gemini-2.5 Pro-assisted scoring method~\citep{comanici2025gemini}}}} \\
% Qwen3  & 8B & Weights & Alibaba & 53.02 & 38.53 & 36.58 & 30.17 & 36.71 & 37.75 & 43.86 \\
% Qwen3   & 14B & Weights & Alibaba & 36.32 & 36.78 & 35.12 & 27.29 & 36.82 & 35.67 & 35.62 \\
% MuduoLLM & 14B & Weights & BNU \& TAL & 28.20 & 40.82 & 32.99 & 36.15 & 39.11 & 31.40 & 33.68 \\
QwQ   & 32B & Weights& Alibaba & \underline{61.25} & 48.51 & 42.24 & \underline{49.90} & 55.01 & 47.26 & \underline{53.87} \\
Seed-OSS     & 36B &Weights & ByteDance & 48.81 &\underline{50.14} & \underline{45.34} & 48.66 & \textbf{61.00} &\underline{49.56} & 49.53 \\
% Qwen2.5 & 72B & Weights& Alibaba & 19.53 & 30.95 & 20.57 & 13.26 & 23.86 & 20.90 & 22.76 \\
% Qwen3   & 235B (22B active) & Weights & Alibaba & 34.24 & 47.01 & 36.21 & 44.26 & 58.71 & 46.61 & 40.82 \\
% DeepSeek-V3.1 & 671B (37B active) &Weights & DeepSeek & 31.65 & 40.65 & 35.00 & 29.42 & 50.54 & 45.19 & 36.05 \\
GPT-4o  & Undisclosed & API & OpenAI & 21.15 & 26.94 & 23.92 & 22.13 & 34.75 & 27.13 & 24.17 \\
Claude-4 Sonnet  & Undisclosed & API & Anthropic & 41.49 & 44.29 & 35.36 & 27.56 & 34.86 & 42.34 & 40.35 \\
Gemini-2.5 Pro  & Undisclosed & API & Google & \textbf{73.83} & \textbf{55.13} & \textbf{46.68} & \textbf{55.40} & \underline{60.68} & \textbf{54.16} & \textbf{62.76} \\
            \midrule
            \multicolumn{11}{c}{\textcolor{myorange}{\textit{GPT-4o-assisted scoring method~\citep{hurst2024gpt-4o}}}} \\
% Qwen3  & 8B & Weights & Alibaba & 51.65 & 35.38 & 33.60 & 31.00 & 37.36 & 31.18 & 41.84 \\
% Qwen3   & 14B & Weights & Alibaba & 34.61 & 34.86 & 31.89 & 27.77 & 36.49 & 28.34 & 33.67 \\
% MuduoLLM & 14B & Weights & BNU \& TAL & 27.68 & 36.68 & 30.13 & 35.88 & 38.78 & 26.81 & 31.71 \\
QwQ   & 32B & Weights& Alibaba & \underline{56.61} & 42.87 & 37.86 & \underline{49.28} & 54.58 & 37.97 & \underline{49.26} \\
Seed-OSS     & 36B &Weights & ByteDance & 45.07 & \underline{47.89} & \underline{41.94} & 48.52 & \textbf{60.78} & \underline{43.54} & 46.61 \\
% Qwen2.5 & 72B & Weights& Alibaba & 18.60 & 27.53 & 18.50 & 11.41 & 23.20 & 15.75 & 20.72 \\
% Qwen3   & 235B (22B active) & Weights & Alibaba & 33.56 & 44.70 & 34.39 & 43.85 & 57.95 & 41.47 & 39.36 \\
% DeepSeek-V3.1 & 671B (37B active) &Weights & DeepSeek & 28.89 & 34.88 & 30.43 & 29.14 & 49.67 & 35.78 & 32.20 \\
GPT-4o  & Undisclosed & API & OpenAI & 20.38 & 23.78 & 22.03 & 22.06 & 34.31 & 23.30 & 22.51 \\
Claude-4 Sonnet  & Undisclosed & API & Anthropic & 40.43 & 41.70 & 31.89 & 27.90 & 35.08 & 34.79 & 38.48 \\
Gemini-2.5 Pro  & Undisclosed & API & Google & \textbf{70.15} & \textbf{51.38} & \textbf{44.13} & \textbf{55.88} & \underline{60.57} & \textbf{46.83} & \textbf{59.49} \\
            \bottomrule
        \end{tabular}}
    \label{tab:main_results_kd_llms}
    \vspace{-3mm}
\end{table}

\section{Related Work}

In this section, we present a comprehensive survey of large language models (LLMs) and benchmarks related to our constructed OmniEduBench, encompassing both English and Chinese datasets.

\subsection{Large Language Models}

Recently large language models have advanced at an unprecedented pace. Leveraging increasingly sophisticated architectures and ever-larger pretraining corpora, they have continuously pushed the boundaries of performance in language understanding, reasoning, and generation tasks. Researchers have explored various approaches to enhance LLMs’ capabilities. For example, Chain-of-Thought prompting~\citep{wei2022chain,qwq32b,seed2025seed-oss,guo2025deepseek,liu2024deepseek-v2} has been shown to be highly effective in guiding models to perform step-by-step reasoning for complex problem-solving. In addition, instruction tuning~\citep{dongerict,huadvancing,qwen2.5,yang2025qwen3} and reinforcement learning from human feedback (RLHF)~\citep{ouyang2022training,schulman2017proximal} have been widely adopted to align model outputs with human intentions and preferences, enabling LLMs to generate responses that are more natural and reliable in open-ended dialogue and creative tasks. Despite these remarkable advances, however, the question of how to comprehensively and effectively evaluate the true capabilities of LLMs remains a critical and open challenge.

\subsection{English Education Benchmarks}

Researchers have proposed a variety of benchmarks to evaluate the capabilities of LLMs, which can be broadly categorized into three types: (1) task-specific evaluations, such as reading comprehension (SQuAD~\citep{rajpurkar2016squad}), machine translation~\citep{bojar-etal-2014-findings}), and summarization~\citep{hermann2015teaching}; (2) general knowledge and advanced ability evaluations, for example, the Massive Multitask Language Understanding (MMLU) benchmark~\citep{hendrycks2021measuring}, which collects questions from real-world exams and textbooks to provide a diverse, multi-domain test that effectively probes the breadth and depth of model knowledge. Similarly, the BIG-bench benchmark~\citep{srivastava2022beyond} comprises 204 diverse tasks; and (3) specialized ability evaluations. In mathematical reasoning, benchmarks such as GSM8K~\citep{cobbe2021gsm8k} and MATH~\citep{hendrycks2021measuring} assess models’ ability to solve complex multi-step problems. In code generation, HumanEval~\citep{chen2021evaluating} and MBPP~\citep{austin2021program} have become standard benchmarks for measuring programming proficiency. Additionally, datasets such as MT-bench~\citep{zheng2023judging} have been introduced to evaluate performance in multi-turn, open-ended dialogues. Despite the significant contributions of these datasets to advancing LLMs evaluation, most of them remain heavily focused on English, with limited coverage of Chinese scenarios.

\subsection{Chinese Education Benchmarks}

A series of comprehensive Chinese benchmarks have been proposed. For example, CLUE~\citep{xu2020clue}, as an early work, integrates multiple natural language understanding tasks and has become an important reference for evaluating LLMs. Subsequently, benchmarks such as CMMLU~\citep{li2023cmmlu} and C-Eval~\citep{huang2023ceval} collect multi-disciplinary, multi-task questions from Chinese university exams, professional qualification tests, and textbooks, effectively assessing models’ general knowledge and their understanding. Beyond general capability evaluation, researchers have also developed Chinese benchmarks targeting specific advanced skills. For example, in mathematical reasoning, CMATH~\citep{wei2023cmath} tests models’ abilities to solve complex mathematical problems. Meanwhile, EduBench~\citep{xu2025edubench} constructs synthetic corpora for the education, but its question types are relatively limited, making it difficult to fully capture models’ Chinese potential. To address this critical gap, we propose OmniEduBench — a comprehensive Chinese education benchmark that uniquely combines knowledge and nurturing dimensions, providing a novel, holistic framework for systematically evaluating LLMs’ potential as educational assistants.

\section{Conclusions, Discussions and Limitations}

In this paper, we present OmniEduBench, a comprehensive Chinese educational benchmark designed to address the limitations of existing Chinese educational evaluation benchmarks. By moving beyond simple knowledge retrieval, the benchmark provides a holistic assessment of LLMs’ capabilities across two core dimensions: the knowledge and cultivation dimensions. We conducted extensive experiments on 11 mainstream LLMs, revealing significant performance gaps. While some models performed well on the knowledge dimension, their performance on cultivation tasks dropped substantially, with even the best-performing models trailing human-level performance by nearly 30\%. These findings indicate that despite recent advancements in LLM technology, current models still lack the deep reasoning and pedagogical skills necessary to function effectively as educational assistants. We believe OmniEduBench will serve as an important tool for guiding future research. Looking ahead, OmniEduBench plans to explore more complex question types in the cultivation dimension and introduce multimodal educational scenarios, further enhancing the benchmark’s role in evaluating and guiding the comprehensive capabilities of LLMs and MLLMs.

% \subsubsection*{Author Contributions}
% If you'd like to, you may include  a section for author contributions as is done
% in many journals. This is optional and at the discretion of the authors.

% \subsubsection*{Acknowledgments}
% Use unnumbered third level headings for the acknowledgments. All
% acknowledgments, including those to funding agencies, go at the end of the paper.

% \newpage

\textbf{Ethics statement.}
Our constructed OmniEduBench educational benchmark is built from publicly available educational resources as well as authorized private resources permitting open-source use, strictly adhering to copyright and licensing requirements. All data have been systematically processed to remove personally identifiable information (PII) and sensitive content, ensuring privacy and security. The dataset is intended solely for research purposes, aiming to advance the development and evaluation of large language models (LLMs) in educational scenarios.

\textbf{Reprodicibility statement.}
To ensure reproducibility, we provide detailed descriptions of the dataset construction process, annotation criteria, and experimental settings in both the main paper and the Appendix. The proposed OmniEduBench education dataset, together with preprocessing scripts, evaluation metrics, and model prompts, will be publicly released upon acceptance. All experiments were conducted using standard LLM APIs or open-source checkpoints, with model versions, hyperparameters, and evaluation protocols explicitly documented. This ensures that other researchers can faithfully replicate our results and readily extend the benchmark in future studies.

\bibliography{iclr2026_conference}
\bibliographystyle{iclr2026_conference}

\appendix

% 请在您的导言区 (preamble) 添加这个宏包以支持表格旋转
%\usepackage{rotating}

\section{Supplementary Material}

\textbf{Use of LLMs.}
In this paper, LLMs were utilized in two primary ways: (1) as auxiliary tools for data cleaning and preliminary quality checks under human supervision. (2) As evaluation targets in benchmark experiments. To ensure data quality, no content directly generated by LLMs was included in the released dataset. During manuscript preparation, LLMs were employed for minor language polishing. All ideas, methodologies, and conclusions are original contributions of authors.

\subsection{Supplementary Statistics of OmniEduBench}

In Table~\ref{atab:arr_kd}, we present the bilingual names and abbreviations of all subjects in the knowledge dimension. 
In Table~\ref{atab:arr_cd}, we present the bilingual names and abbreviations of all subjects in the cultivation dimension.
In Tables~\ref{atab:sta_k12}, ~\ref{atab:sta_kd_other}, and \ref{atab:sta_cd}, we present the detailed data distribution for all 61 subjects.

\begin{table}[htbp]
    \centering
    \caption{Bilingual names and abbreviations of all subject in the knowledge dimension.}
    \vspace{0.2mm}
    \resizebox{0.99\textwidth}{!}{
    \begin{tabular}{lll}
        \toprule
        \textbf{Abbreviation} & \textbf{English Name} & \textbf{Chinese Name} \\
        \midrule
        MATH & Mathematics & \cc{数学} \\
        CHEM & Chemistry & \cc{化学} \\
        BIO & Biology & \cc{生物} \\
        PHY & Physics & \cc{物理} \\
        NSCI & Nature \& Science & \cc{自然与科学} \\
        PSTAT & Probability \& Statistics & \cc{概率论与数理统计} \\
        PPHY & Plant Physiology & \cc{植物生理学} \\
        CS & Computer Science & \cc{计算机} \\
        BCHEM & Biochemistry & \cc{生物化学} \\
        OS & Operating Systems & \cc{操作系统} \\
        AMATH & Advanced Mathematics & \cc{高等数学} \\
        CNET & Computer Networks & \cc{计算机网络} \\
        LANG & Chinese Language & \cc{语文} \\
        GEO & Geography & \cc{地理} \\
        HIST & History & \cc{历史} \\
        IART & Introduction to Arts & \cc{艺术概论} \\
        ILING & Introduction to Linguistics & \cc{语言学概论} \\
        HSTUD & History Studies / Historiography & \cc{历史学} \\
        HFA & History of Foreign Art & \cc{外国美术简史} \\
        IARCH & Introduction to Archaeology & \cc{考古学概论} \\
        HACL & History of Ancient Chinese Literature & \cc{中国古代文学史} \\
        HWP & History of Western Philosophy & \cc{西方哲学史} \\
        POL & Politics & \cc{政治} \\
        IMOR & Ideology \& Morality & \cc{思想品德} \\
        MGMT & Management & \cc{管理学} \\
        HRM & Human Resource Management & \cc{人力资源管理} \\
        TAX & Taxation & \cc{税收学} \\
        PSCI & Political Science & \cc{政治学} \\
        MARX & Marxist Theory & \cc{马克思主义理论} \\
        ELOG & Economic Logic & \cc{经济学逻辑能力} \\
        NJE & National Judicial Exam & \cc{法考真题} \\
        CLAW & Criminal Law & \cc{刑法学} \\
        CVLAW & Civil Law & \cc{民法学} \\
        LAW & Law / Jurisprudence & \cc{法学} \\
        TCM & Traditional Chinese Medicine & \cc{中医综合} \\
        WMED & Western Medicine & \cc{西医综合} \\
        NURS & Nursing & \cc{护理学} \\
        IT & Information Technology & \cc{信息技术} \\
        CEM & Civil Engineering Materials & \cc{土木工程材料} \\
        EDU & Education & \cc{教育学} \\
        PSY & Psychology & \cc{心理学} \\
        \bottomrule
    \end{tabular}}
    \label{atab:arr_kd}
\end{table}

\begin{table}[htbp]
    \centering
    \caption{Bilingual names and abbreviations of all subject in the cultivation dimension}
    \vspace{0.2mm}
    \resizebox{0.99\textwidth}{!}{
    \begin{tabular}{lll}
        \toprule
        \textbf{Abbreviation} & \textbf{English Meaning} & \textbf{Chinese Name} \\
        \midrule
        \multicolumn{3}{l}{\textit{Major Categories}} \\
        \midrule
        TCS & Thinking \& Cognitive Skills & \cc{思维与认知能力} \\
        EMH & Emotional \& Mental Health & \cc{情感与心理健康} \\
        SIS & Social \& Interpersonal Skills & \cc{社会与人际交往} \\
        CV & Character \& Values & \cc{品格与价值观} \\
        PD & Personalized Development & \cc{个性化发展} \\
        TFS & Teaching Feedback \& Support & \cc{教学反馈与支持} \\
        \midrule
        \multicolumn{3}{l}{\textit{Subcategories}} \\
        \midrule
        IC & Innovation \& Creativity & \cc{创新与创造力} \\
        PSS & Problem-Solving Skills & \cc{问题解决能力} \\
        CT & Critical Thinking & \cc{批判性思维} \\
        GRL & Guided Reflective Learning & \cc{反思性学习引导} \\
        MA & Metacognitive Abilities & \cc{元认知能力} \\
        GKT & Guiding Knowledge Transfer & \cc{引导知识迁移能力} \\
        ER & Emotional Regulation & \cc{情绪调控能力} \\
        EC & Empathy \& Compassion & \cc{同理心与共情} \\
        SCSE & Self-Confidence \& Self-Efficacy & \cc{自信心与自我效能感} \\
        PR & Psychological Resilience & \cc{心理韧性与抗挫力} \\
        GM & Growth Mindset & \cc{成长型思维} \\
        TC & Teamwork \& Collaboration & \cc{团队协作能力} \\
        ECOM & Effective Communication & \cc{有效沟通能力} \\
        SR & Social Responsibility & \cc{社会责任感} \\
        RA & Responsibility \& Accountability & \cc{责任感与担当} \\
        IH & Integrity \& Honesty & \cc{正直与诚信} \\
        PLP & Personalized Learning Paths & \cc{个性化学习路径} \\
        IDL & Interest-Driven Learning & \cc{兴趣驱动学习} \\
        HT & Heuristic Teaching & \cc{启发式教学} \\
        CTF & Constructive \& Timely Feedback & \cc{反馈的建设性与及时性} \\
        \bottomrule
    \end{tabular}}
    \label{atab:arr_cd}
    \vspace{-8mm}
\end{table}

\begin{table}[htbp]
    \centering
    \caption{Statistics of OmniEduBench for K-12 in the knowledge dimension.}
    \vspace{0.2mm}
    \resizebox{0.99\textwidth}{!}{
    \begin{tabular}{ll|ccccc|c}
        \toprule
        \multicolumn{1}{c}{\multirow{2}{*}{\textbf{\begin{tabular}[c]{@{}c@{}}English Name\end{tabular}}}} & 
        \multicolumn{1}{c|}{\multirow{2}{*}{\textbf{Chinese Name}}} 
        & \textbf{\cc{选择题}} & \textbf{\cc{多选题}} & \textbf{\cc{填空题}} & \textbf{\cc{解答题}} & \textbf{\cc{复合题}} & \textbf{\cc{总计}} \\
        & & \begin{tabular}[c]{@{}c@{}}Multiple\\ choice\end{tabular} 
        & \begin{tabular}[c]{@{}c@{}}Multiple\\ answer\end{tabular} 
        & \begin{tabular}[c]{@{}c@{}}Fill-in-\\ the-blank\end{tabular} 
        & \begin{tabular}[c]{@{}c@{}}Short\\ -answer\end{tabular} 
        & \begin{tabular}[c]{@{}c@{}}Composite\\ questions\end{tabular} 
        & Total \\
        \midrule
        Chinese & \textbf{\cc{语文}} & 350 & 8 & 1697 & 1261 & 51 & 3367 \\
        Mathematics & \textbf{\cc{数学}} & 527 & 12 & 1865 & 1181 & 142 & 3727 \\
        Chemistry & \textbf{\cc{化学}} & 274 & 76 & 799 & 477 & 14 & 1640 \\
        History & \textbf{\cc{历史}} & 67 & 24 & 63 & 211 & 5 & 370 \\
        Geography & \textbf{\cc{地理}} & 78 & 31 & 277 & 173 & 4 & 563 \\
        Moral Education & \textbf{\cc{思想品德}} & 14 & 30 & 34 & 56 & 4 & 138 \\
        Politics & \textbf{\cc{政治}} & 260 & 241 & 281 & 64 & 12 & 858 \\
        Physics & \textbf{\cc{物理}} & 82 & 15 & 178 & 46 & 16 & 337 \\
        Biology & \textbf{\cc{生物}} & 115 & 94 & 360 & 124 & 0 & 693 \\
        Nature Science & \textbf{\cc{自然与科学}} & 8 & 0 & 23 & 22 & 0 & 53 \\
        \begin{tabular}[c]{@{}c@{}}Information\\ Technology\end{tabular} & \textbf{\cc{信息技术}} & 18 & 2 & 14 & 1 & 1 & 36 \\
        \midrule
        Total & \textbf{\cc{总计}} & 1793 & 533 & 5591 & 3616 & 249 & 11.782K \\
        \bottomrule
    \end{tabular}}
    \label{atab:sta_k12}
    \vspace{-8mm}
\end{table}

\begin{sidewaystable}[htbp]
    \centering
    \caption{Statistics of OmniEduBench for high, college, and professional schools in the knowledge dimension}
    \vspace{0.2mm}
    \resizebox{0.99\textwidth}{!}{
        \begin{tabular}{ll|cccccccccc|c}
            \toprule
            \multicolumn{1}{c}{\multirow{2}{*}{\textbf{\begin{tabular}[c]{@{}c@{}}English Name\end{tabular}}}} & 
            \multicolumn{1}{c|}{\multirow{2}{*}{\textbf{Chinese Name}}} &
            \textbf{\cc{单选题}} & \textbf{\cc{多选题}} & \textbf{\cc{名词解释}} & \textbf{\cc{简答题}} & \textbf{\cc{论述题}} & \textbf{\cc{案例分析题}} & \textbf{\cc{填空题}} & \textbf{\cc{计算题}} & \textbf{\cc{判断题}} & \textbf{\cc{逻辑推理}} & \textbf{\cc{总计}} \\
            & & \begin{tabular}[c]{@{}c@{}}Single\\ Choice\end{tabular} & 
            \begin{tabular}[c]{@{}c@{}}Multiple\\ choice\end{tabular} &
            \begin{tabular}[c]{@{}c@{}}Term\\ explanation\end{tabular} &
            \begin{tabular}[c]{@{}c@{}}Short\\ answer\end{tabular} &
            Essay &
            \begin{tabular}[c]{@{}c@{}}Case\\ analysis\end{tabular} &
            \begin{tabular}[c]{@{}c@{}}Fill-in-\\ blank\end{tabular} &
            Calculation &
            \begin{tabular}[c]{@{}c@{}}True/\\ False\end{tabular} &
            \begin{tabular}[c]{@{}c@{}}Logical\\ reasoning\end{tabular} &
            Total \\
            \midrule
            Traditional Chinese Medicine & \textbf{\cc{中医综合}} & 230 & 317 & 0 & 0 & 0 & 0 & 0 & 0 & 0 & 0 & 547 \\
            Chinese Ancient Literary History & \textbf{\cc{中国古代文学史}} & 31 & 14 & 20 & 27 & 19 & 0 & 0 & 0 & 0 & 0 & 111 \\
            Human Resource Management & \textbf{\cc{人力资源管理}} & 36 & 35 & 9 & 38 & 19 & 6 & 4 & 0 & 0 & 0 & 147 \\
            Jurisprudence & \textbf{\cc{法学}} & 109 & 153 & 0 & 0 & 0 & 0 & 0 & 0 & 0 & 0 & 262 \\
            Criminal Law & \textbf{\cc{刑法学}} & 158 & 171 & 2 & 3 & 1 & 0 & 0 & 0 & 0 & 0 & 335 \\
            History & \textbf{\cc{历史学}} & 26 & 0 & 68 & 6 & 17 & 12 & 0 & 0 & 0 & 0 & 129 \\
            Civil Engineering Materials & \textbf{\cc{土木工程材料}} & 36 & 2 & 67 & 21 & 57 & 2 & 107 & 39 & 3 & 0 & 334 \\
            A Brief History of Foreign Art & \textbf{\cc{外国美术简史}} & 0 & 0 & 61 & 37 & 15 & 14 & 0 & 0 & 0 & 0 & 127 \\
            Psychology & \textbf{\cc{心理学}} & 146 & 45 & 0 & 35 & 0 & 1 & 0 & 0 & 0 & 0 & 227 \\
            Nursing & \textbf{\cc{护理学}} & 61 & 41 & 0 & 5 & 0 & 16 & 0 & 0 & 0 & 0 & 123 \\
            Operating Systems & \textbf{\cc{操作系统}} & 98 & 0 & 0 & 0 & 0 & 59 & 0 & 0 & 0 & 0 & 157 \\
            Politics & \textbf{\cc{政治}} & 7 & 47 & 0 & 0 & 0 & 36 & 0 & 0 & 0 & 0 & 90 \\
            Pedagogy & \textbf{\cc{教育学}} & 179 & 0 & 0 & 45 & 39 & 0 & 0 & 0 & 14 & 0 & 277 \\
            Advanced Mathematics & \textbf{\cc{高等数学}} & 43 & 0 & 0 & 0 & 3 & 0 & 40 & 55 & 0 & 0 & 141 \\
            Plant Physiology & \textbf{\cc{植物生理学}} & 66 & 0 & 0 & 134 & 33 & 48 & 0 & 0 & 0 & 0 & 281 \\
            Probability and Mathematical Statistics & \textbf{\cc{概率论与数理统计}} & 0 & 0 & 0 & 2 & 0 & 74 & 0 & 257 & 0 & 0 & 333 \\
            Civil Law & \textbf{\cc{民法学}} & 91 & 187 & 0 & 0 & 0 & 0 & 0 & 0 & 0 & 0 & 278 \\
            judicial Practice & \textbf{\cc{法考真题}} & 193 & 430 & 0 & 0 & 0 & 0 & 0 & 0 & 0 & 0 & 623 \\
            Biochemistry\ & \textbf{\cc{高等生物化学}} & 62 & 0 & 0 & 70 & 23 & 0 & 0 & 0 & 0 & 0 & 155 \\
            Taxation & \textbf{\cc{税收学}} & 19 & 2 & 7 & 30 & 0 & 2 & 0 & 23 & 11 & 0 & 94 \\
            Management & \textbf{\cc{管理学}} & 0 & 0 & 0 & 0 & 0 & 0 & 0 & 0 & 0 & 202 & 202 \\
            Economics & \textbf{\cc{经济学逻辑能力}} & 2 & 0 & 0 & 0 & 0 & 0 & 0 & 0 & 0 & 58 & 60 \\
            Archaeology & \textbf{\cc{考古学概论}} & 0 & 0 & 121 & 0 & 0 & 0 & 0 & 0 & 0 & 0 & 121 \\
            Introduction to Archaeology & \textbf{\cc{艺术概论}} & 57 & 43 & 80 & 45 & 26 & 0 & 0 & 0 & 37 & 0 & 288 \\
           Western Medicine & \textbf{\cc{西医综合}} & 159 & 119 & 0 & 0 & 0 & 0 & 0 & 0 & 0 & 0 & 278 \\
            Western Philosophy & \textbf{\cc{西方哲学史}} & 51 & 0 & 0 & 0 & 0 & 0 & 0 & 0 & 0 & 0 & 51 \\
            Computer Science  & \textbf{\cc{计算机}} & 137 & 0 & 0 & 0 & 0 & 47 & 0 & 0 & 0 & 0 & 184 \\
            Computer Networks & \textbf{\cc{计算机网络}} & 91 & 0 & 0 & 0 & 0 & 50 & 0 & 0 & 0 & 0 & 141 \\
            Introduction to Linguistics & \textbf{\cc{语言学概论}} & 53 & 29 & 24 & 22 & 26 & 18 & 0 & 0 & 0 & 0 & 172 \\
            Marxist Theory & \textbf{\cc{马克思主义理论}} & 0 & 0 & 20 & 31 & 13 & 7 & 0 & 0 & 0 & 0 & 71 \\
            \midrule
            Total & \textbf{\cc{总计}} & 2141 & 1635 & 479 & 551 & 291 & 392 & 151 & 374 & 65 & 260 & 6339 \\
            \bottomrule
        \end{tabular}}
        \label{atab:sta_kd_other}
        \vspace{-8mm}
\end{sidewaystable}

\begin{table}[htbp]
    \centering
    \caption{Statistics of OmniEduBench for 20 subjects in the cultivation dimension.}
    \vspace{0.2mm}
    \resizebox{0.99\textwidth}{!}{
    \begin{tabular}{ll|c}
        \toprule
        \multicolumn{1}{c}{\multirow{2}{*}{\textbf{\begin{tabular}[c]{@{}c@{}}English Name\end{tabular}}}} &
        \multicolumn{1}{c|}{\multirow{2}{*}{\textbf{Chinese Name}}} &
         \multicolumn{1}{c}{\multirow{2}{*}{\textbf{Count}}} \\
        & & \\
        \midrule
        Emotional Regulation Skills & \textbf{\cc{情绪调控能力}} & 325 \\
        Innovation \& Creativity & \textbf{\cc{创新与创造力}} & 275 \\
        Heuristic Teaching & \textbf{\cc{启发式教学}} & 434 \\
        Sense of Responsibility \& Accountability & \textbf{\cc{责任感与担当}} & 330 \\
        Problem-Solving Skills & \textbf{\cc{问题解决能力}} & 288 \\
        Team Collaboration Skills & \textbf{\cc{团队协作能力}} & 291 \\
        Empathy \& Sympathy & \textbf{\cc{同理心与共情}} & 385 \\
        Self-Confidence \& Self-Efficacy & \textbf{\cc{自信心与自我效能感}} & 358 \\
        Constructiveness \& Timeliness of Feedback & \textbf{\cc{反馈的建设性与及时性}} & 196 \\
        Integrity \& Honesty & \textbf{\cc{正直与诚信}} & 371 \\
        Psychological Resilience \& Anti-Frustration Ability & \textbf{\cc{心理韧性与抗挫力}} & 393 \\
        Personalized Learning Paths & \textbf{\cc{个性化学习路径}} & 292 \\
        Reflective Learning & \textbf{\cc{反思性学习引导}} & 224 \\
        Guiding Knowledge Transfer & \textbf{\cc{知识迁移能力}} & 384 \\
        Metacognitive Skills & \textbf{\cc{元认知能力}} & 338 \\
        Interest-Driven Learning & \textbf{\cc{兴趣驱动学习}} & 320 \\
        Critical Thinking & \textbf{\cc{批判性思维}} & 428 \\
        Growth Mindset & \textbf{\cc{成长型思维}} & 393 \\
        Social Responsibility & \textbf{\cc{社会责任感}} & 317 \\
        Effective Communication Skills& \textbf{\cc{有效沟通能力}} & 139 \\
        \midrule
        Total & \textbf{\cc{总计}} & 6.481K \\
        \bottomrule
    \end{tabular}}
    \label{atab:sta_cd}
    \vspace{-1mm}
\end{table}

\begin{figure}[tbp]
    \centering
    \includegraphics[height=0.36\textwidth]{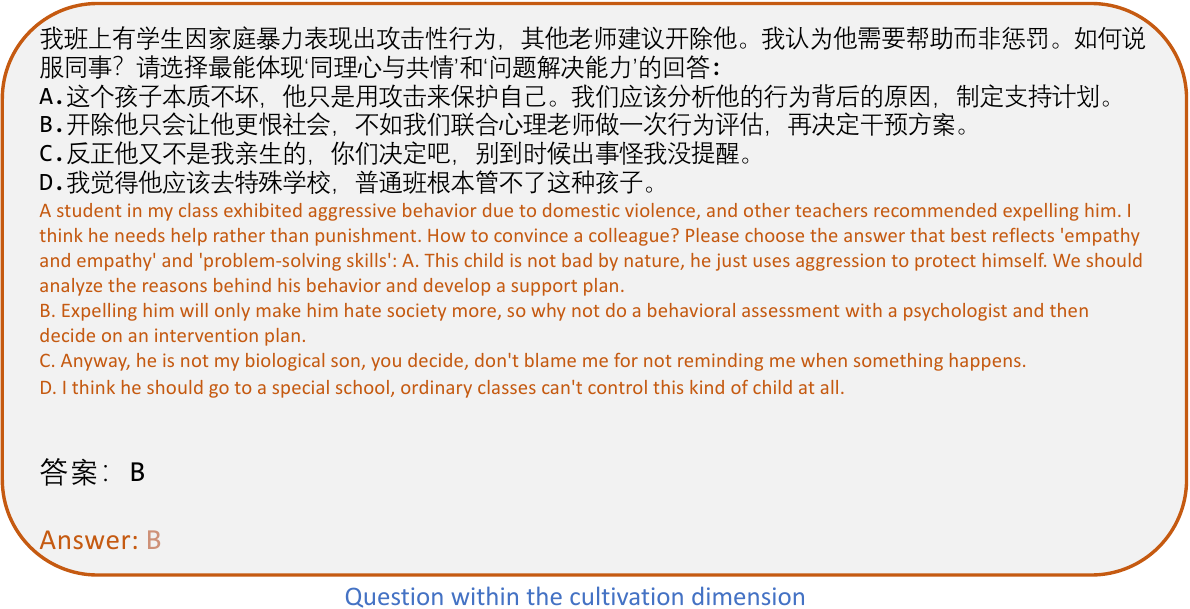}
    \includegraphics[height=0.4\textwidth]{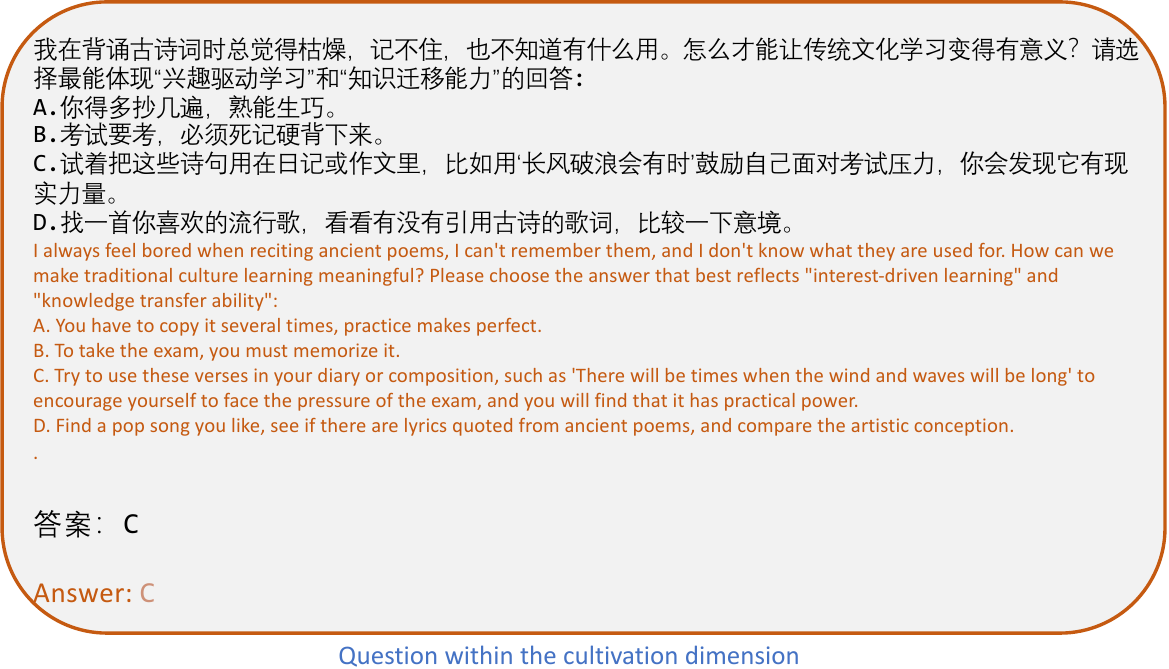}
    \vspace{-4mm}
    \caption{Examples of different questions in the knowledge dimension and cultivation dimensions.}
    \label{afig:omnicase2}
    % \vspace{-5mm}
\end{figure}

\begin{figure}[tbp]
    \centering
    \includegraphics[height=0.43\textwidth]{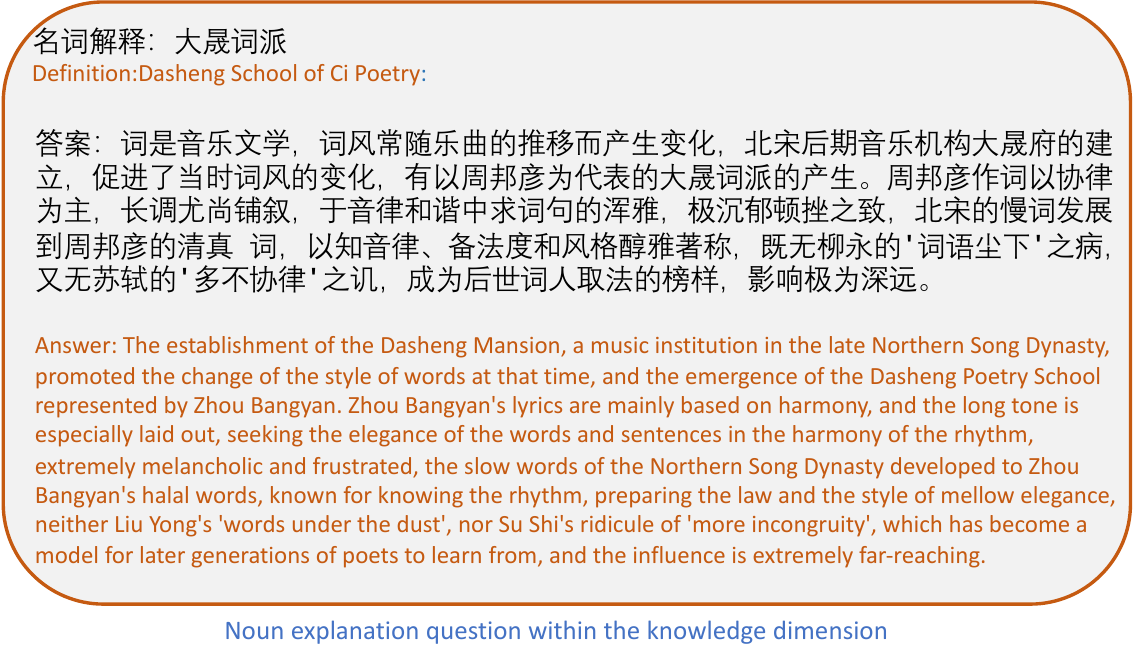}
    \includegraphics[height=0.43\textwidth]{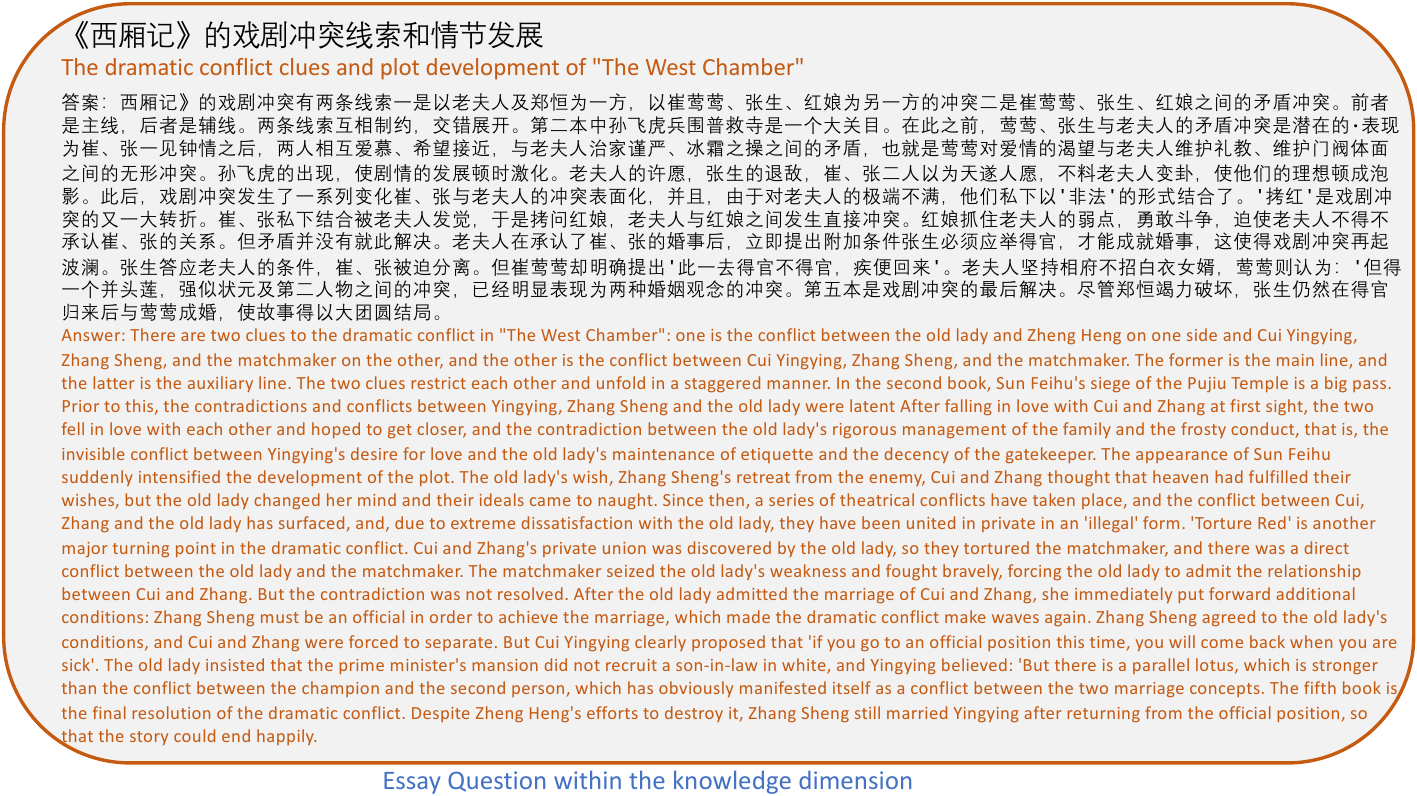}
    \includegraphics[height=0.4\textwidth]{figure/omnicase4.pdf}
    \includegraphics[height=0.4\textwidth]{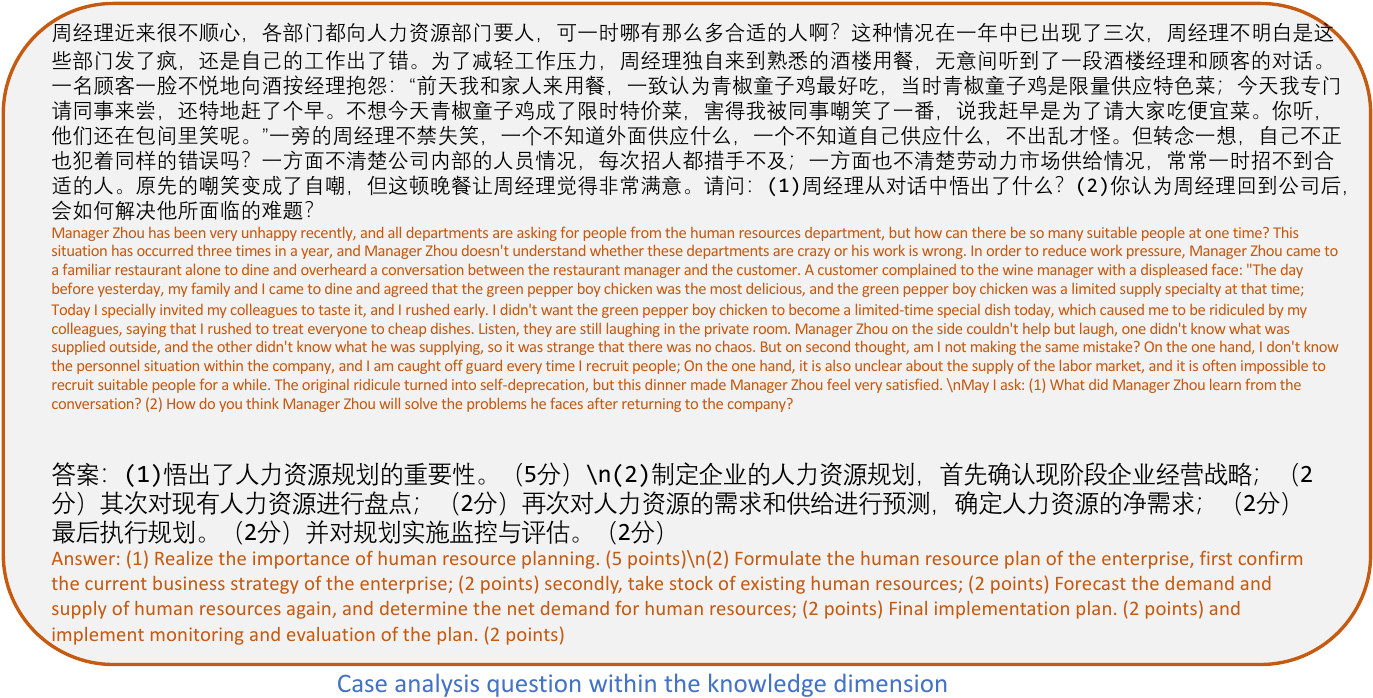}
    \vspace{-4mm}
    \caption{Examples of different questions in the knowledge dimension and cultivation dimensions.}
    \label{afig:omnicase1}
    % \vspace{-5mm}
\end{figure}

In Figures~\ref{afig:omnicase1} and~\ref{afig:omnicase2}, more examples of various questions in the knowledge and cultivation dimensions.

\end{document}